\documentclass{article}

\usepackage[preprint]{neurips_2026}

\usepackage[utf8]{inputenc}
\usepackage[T1]{fontenc}
\usepackage{booktabs}
\usepackage{amsfonts}
\usepackage{microtype}
\usepackage{xcolor}
\usepackage{graphicx}
\graphicspath{{figures/}}
\usepackage{wrapfig}
\usepackage{multirow}
\usepackage{amssymb}
\usepackage{pifont}
\usepackage{amsmath}
\usepackage{hyperref}
\usepackage{url}
\usepackage{natbib}
\usepackage{algorithm}
\usepackage{algorithmic}
\usepackage{xspace}
\usepackage{makecell}
\usepackage{colortbl}
\usepackage[capitalize,noabbrev]{cleveref}

\definecolor{blond}{rgb}{0.98, 0.94, 0.75}

\newcommand{\ours}{CMKL\xspace}
\newcommand{\fullname}{Continual Multimodal Knowledge Graph Learner\xspace}

\title{\ours: Modality-Aware Continual Learning \\
for Evolving Biomedical Knowledge Graphs}

\author{%
  Yousef A. Radwan\\
  Technology, Innovation, Entrepreneurship Department \\
  King Abdullah University of Science and Technology\\
  \texttt{yousef.radwan@kaust.edu.sa}
  \And
  Yao Li\\
  School of Computing and Information Systems\\
  The University of Melbourne\\
  \texttt{yao.li5@student.unimelb.edu.au} \\
  \And
  Qing Qing\\
  College of Computer Science and Technology\\
  Jilin University\\
  \texttt{qingqing25@mails.jlu.edu.cn} \\
  \And
  Ziqi Xu\\
   School of Computing Technologies \\
   RMIT University \\
  \texttt{ziqi.xu@rmit.edu.au} \\
  \And
  Qixin Zhang\\
  College of Computing and Data Science\\
  Nanyang Technological University\\
  \texttt{qixin.zhang@ntu.edu.sg} \\
  \And
  Yongcheng Jing\\
  College of Computing \& Data Science\\
  Nanyang Technological University\\
  \texttt{yongcheng.jing@ntu.edu.sg} \\
  \And
  Renqiang Luo\thanks{Corresponding authors.} \\
  College of Computer Science and Technology\\
  Jilin University\\
  \texttt{lrenqiang@jlu.edu.cn} \\
   \And
   Xikun Zhang\footnotemark[1] \\
   School of Computing Technologies \\
   RMIT University \\
   \texttt{xikun.zhang@rmit.edu.au} \\   
}

\begin{document}

\maketitle

\begin{abstract}
Biomedical knowledge graphs (KGs) are increasingly large, dynamic, and multimodal, driven by rapid advances in biotechnology such as high-throughput sequencing. Machine learning models can infer previously unobserved biomedical relationships and characterize biomedical entities in these graphs, but existing knowledge graph embedding (KGE) methods and their continual learning (CL) extensions either assume static graph structure or fail to exploit multimodal information under evolving data distributions. They also apply uniform regularization across all model parameters, ignoring that different modalities may exhibit distinct forgetting dynamics as the graph evolves.
We propose the Continual Multimodal Knowledge Graph Learner (CMKL), a CL framework for biomedical KGs that natively encodes structure (R-GCN), text (frozen BiomedBERT), and molecules (Morgan fingerprints), fuses them through a Mixture-of-Experts (MoE) router, and protects previously learned knowledge with standard EWC regularization and a K-means-diverse multimodal replay buffer.
We evaluate CMKL on a 129K-entity biomedical continual benchmark with 10 tasks. On continual biomedical entity classification, CMKL reaches AP $0.591$ versus $0.370$ for the strongest structural baseline, a $+60\%$ gain that is driven by access to multimodal features and preserved across the sequence with near-zero forgetting (AF $0.008$). On continual relationship prediction, CMKL reaches AP $0.062$, matching Naive Sequential and EWC ($0.058$) within seed noise and outperforming Joint Training ($0.047$, $p{=}0.045$) and LKGE ($0.039$). A frozen-text ablation reaches AP $0.136$, more than double any jointly trained model, yet that signal is unreachable by margin-ranking gradients: the greedy-modality asymmetry lives at the \emph{representation level}, not the fusion level, and MoE routing manages it by suppressing the unreachable modality without forcing it through a learned bottleneck. Our code is available at \url{https://github.com/yradwan147/cmkl-neurips2026}.
\end{abstract}

\section{Introduction}
\label{sec:intro}
Modern biomedical knowledge graphs (KGs) integrate heterogeneous modalities, including relational structure encoding entity interactions, textual descriptions from medical ontologies, and molecular representations from chemical databases, into unified resources that power drug discovery, disease understanding, and target identification~\citep{chandak2023building, huang2024foundation}. These modalities provide complementary signals: a drug's molecular fingerprint constrains its binding targets, its textual description captures clinical context, and its graph neighborhood reveals functional associations. As biomedical databases are continuously updated, KG embedding models must adapt to new knowledge without forgetting previously learned associations, a challenge known as \emph{catastrophic forgetting}~\citep{kirkpatrick2017overcoming}.

Existing continual knowledge graph embedding (CKGE) methods, including EWC-based regularization~\citep{kirkpatrick2017overcoming}, experience replay~\citep{rolnick2019experience}, and the LKGE framework~\citep{cui2023lifelong}, address forgetting exclusively through the lens of graph structure. They treat entities as featureless identifiers and apply uniform regularization across all parameters, implicitly assuming that all components forget at the same rate. This assumption ignores the multimodal nature of biomedical entities: textual descriptions, molecular fingerprints, and structural topology may each experience different distribution shifts as the graph evolves, yet no existing CKGE method leverages or protects these distinct modalities.

Building a continual multimodal KG learner raises a question that single-modality CKGE does not face: what should ``not forgetting'' mean when one encoder is frozen and stable, another is trainable and topologically drifting, and a third sees only a small fraction of entities? We propose \fullname (\ours), which addresses this question with two coupled mechanisms: a Mixture-of-Experts (MoE) fusion module whose router can suppress an unreachable modality without forcing it through a bottleneck, and a K-means-diverse multimodal replay buffer that revisits topologically distinct exemplars with all their modality features intact. These are paired with a standard EWC regularizer applied per encoder as a configuration choice; the per-encoder $\lambda$ ablation is reported in \cref{subsec:cl_ablation} and matches a uniform $\lambda$ on the decoder we evaluate.

A surprising finding emerges from this design. A frozen BiomedBERT encoder, exposed to no link-prediction signal, posts a filtered MRR roughly twice that of any jointly trained multimodal model on this benchmark. The cause is not that text is ``better'' than structure; it is that frozen pretrained semantics produces a useful evaluation-time clustering that margin-ranking loss cannot create or improve. Any fusion that backpropagates through both encoders ends up suppressing the frozen one, because the trainable encoder reduces the loss faster. This is a representation-level instance of the greedy-modality phenomenon~\citep{wu2022characterizing}, not a fusion-level one, and it dictates how a multimodal CKGE system should be built. We confirm this through a systematic comparison of fusion strategies, namely MoE routing, gated cross-attention, concatenation, and score-level fusion with OGM-GE gradient modulation~\citep{peng2022balanced}: embedding-level fusions cluster within seed noise (AP $0.060$--$0.063$) while score-level fusion falls to $0.049$; we adopt MoE primarily because its learned router exposes interpretable per-entity modality weights, which makes the encoder-level asymmetry diagnosable.

We evaluate \ours on a 129K-entity, 8.1M-edge biomedical continual learning benchmark with 10 tasks derived from real-world database updates. On continual biomedical entity classification, a decoder-agnostic test of whether multimodal information transfers across tasks, \ours reaches AP\,=\,$0.591$, a $60\%$ absolute gain over the strongest baseline (Joint Training, $0.370$), with near-zero forgetting. On continual relationship prediction, \ours reaches AP\,=\,$0.062 \pm 0.010$ filtered Mean Reciprocal Rank (MRR) with DistMult, a $7\%$ same-decoder gain over EWC, $1.3\times$ over Joint Training ($0.047$), and $1.6\times$ over LKGE ($0.039$); the headroom is bounded by a decoder-geometry constraint analyzed in \cref{subsec:analysis}.

Our contributions are: (i) \ours, a continual KG learner that natively encodes structure, text, and molecules and combines MoE fusion with K-means-diverse multimodal replay; (ii) a representation-level diagnosis of the greedy-modality problem in multimodal KG learning, explaining why a frozen text encoder beats any trainable fusion on filtered MRR yet contributes nothing the trainable system can retain; and (iii) comprehensive empirical results on a 129K-entity biomedical benchmark: $+60\%$ AP on biomedical entity classification preserved across the continual sequence with near-zero forgetting (the gap is feature-driven, and \ours's contribution is preserving it across tasks rather than producing it), and link-prediction results that match the strongest structural CL methods within seed noise (AP $0.062$ vs $0.058$) while exhibiting markedly different per-task and per-stratum behavior. Per-encoder EWC was evaluated but matches uniform EWC on the bilinear decoder we report; we discuss it as a configuration choice, not a methodological contribution.

\section{Related Work}
\label{sec:related}
\paragraph{Continual Knowledge Graph Embedding.}
LKGE~\citep{cui2023lifelong} provides a modular framework combining KGE models (TransE, DistMult, RotatE) with continual strategies (EWC, experience replay, distillation); IncDE~\citep{liu2024incde} handles incremental KG embedding via incremental distillation; Daruna et al.~\citep{daruna2021continual} study continual graph learning for robotics. Concurrent work by Li et al.~\citep{li2026mrckg} proposes MRCKG, combining curriculum learning and contrastive replay for continual multimodal KGE on general-domain KGs ($\leq$15K entities) with synthetic temporal splits. In contrast, \ours targets real biomedical KG evolution at 129K+ entity scale with domain-specific molecular modalities, per-modality Fisher regularization (vs.\ entity-level preservation), and a systematic analysis of the greedy modality problem. Prior to MRCKG, all CKGE methods operated exclusively on graph structure and applied uniform regularization, ignoring the multimodal attributes in modern biomedical KGs.

\paragraph{Multimodal Knowledge Graph Embedding.}
MoMoK~\citep{zhang2025momok} introduces modality-aware KG completion using \emph{relation-guided} MoE routing with expert disentanglement via mutual information minimization; our MoE design uses \emph{entity-level} routing, better suited to biomedical KGs where modality availability is determined by entity type (e.g., only drugs have molecular fingerprints). NativE~\citep{zhang2024native} and AdaMF-MAT~\citep{zhang2024adamf} tackle modality imbalance via relation-guided dual adaptive fusion and adaptive modality weights respectively. MCLEA~\citep{lin2022mclea} employs contrastive learning for multimodal entity alignment. MoSE~\citep{zhao2022mose} proposes score-level fusion with per-modality decoders. IMF~\citep{Zheng2023IMF} uses interaction-based fusion, and OTKGE~\citep{Cao2022OTKGE} applies optimal transport to align modality-specific embedding spaces. Liang et al.~\citep{liang2024foundations} survey fusion strategies (early, late, hybrid). All these methods assume a static graph setting and do not address the catastrophic forgetting issue.

\paragraph{Greedy Modality and Multimodal Optimization.}
The \emph{greedy modality} problem~\citep{wu2022characterizing} arises when one modality converges faster during joint training, suppressing slower modalities. OGM-GE~\citep{peng2022balanced} addresses this through gradient modulation; CGGM~\citep{guo2024cggm} extends it by considering gradient magnitude and direction. These gradient-level solutions operate during optimization; our MoE routing instead operates at the \emph{representation level}, allowing the router to suppress unhelpful modalities without gradient surgery. The problem is particularly acute when mixing frozen pretrained encoders (e.g., BiomedBERT) with jointly trained components. Our work provides the first systematic study of this phenomenon in the continual KG embedding setting.

\paragraph{Biomedical KG Reasoning and Continual Learning Mechanisms.}
PrimeKG~\citep{chandak2023building} integrates 20+ biomedical databases; TxGNN~\citep{huang2024foundation} leverages it for zero-shot drug repurposing. Both treat the KG as static. EWC~\citep{kirkpatrick2017overcoming} prevents forgetting by penalizing changes to important parameters via the Fisher information; Synaptic Intelligence~\citep{zenke2017continual} accumulates importance online; replay-based methods~\citep{rolnick2019experience, lopez2017gradient, chaudhry2019efficient} revisit stored examples. Standard EWC computes a single Fisher matrix over all parameters; we apply it with per-encoder $\lambda$ values as a configuration choice (without claiming this as a methodological contribution; an ablation against uniform $\lambda$ is in \cref{subsec:cl_ablation}).

\textbf{\ours} combines three complementary mechanisms that are individually known but, to our knowledge, have not been studied together in the continual multimodal biomedical-KG setting: per-modality Fisher regularization, K-means-diverse multimodal memory replay, and entity-level MoE routing. The empirical question this paper asks is which of these mechanisms actually matter on a real biomedical KG: we find MoE fusion is load-bearing for managing the frozen-vs-learnable encoder asymmetry, replay is the dominant forgetting-mitigation mechanism at our scale, and per-modality $\lambda$ in EWC matches uniform $\lambda$ on the bilinear DistMult decoder we report (\cref{subsec:cl_ablation}).

\section{Methodology}
\label{sec:method}

CMKL is designed around an asymmetry between the three modalities of biomedical KGs: graph topology drifts whenever a new PrimeKG release adds or rewires edges, a drug's Morgan fingerprint is fixed once computed, and a frozen BiomedBERT description has no gradient at all. Formally, we observe an evolving biomedical knowledge graph as a sequence of snapshots $\{G_1, G_2, \ldots, G_T\}$, where each $G_t = (\mathcal{V}_t, \mathcal{E}_t, \mathcal{R}_t, \mathbf{X}_t)$ consists of entities $\mathcal{V}_t$, typed edges $\mathcal{E}_t \subseteq \mathcal{V}_t \times \mathcal{R}_t \times \mathcal{V}_t$, relation types $\mathcal{R}_t$, and multimodal entity attributes $\mathbf{X}_t$. The benchmark defines a task sequence $\mathcal{T} = \{T_1, \ldots, T_K\}$ via entity-type grouping, where each task $T_k$ provides training triples for a specific entity-type group. The objective is to learn entity embeddings $\mathbf{h}_v \in \mathbb{R}^D$ that score link prediction across all tasks without catastrophic forgetting.

\subsection{Architecture Overview}
\label{subsec:overview}

\begin{figure*}[t]
\centering
\includegraphics[width=\textwidth]{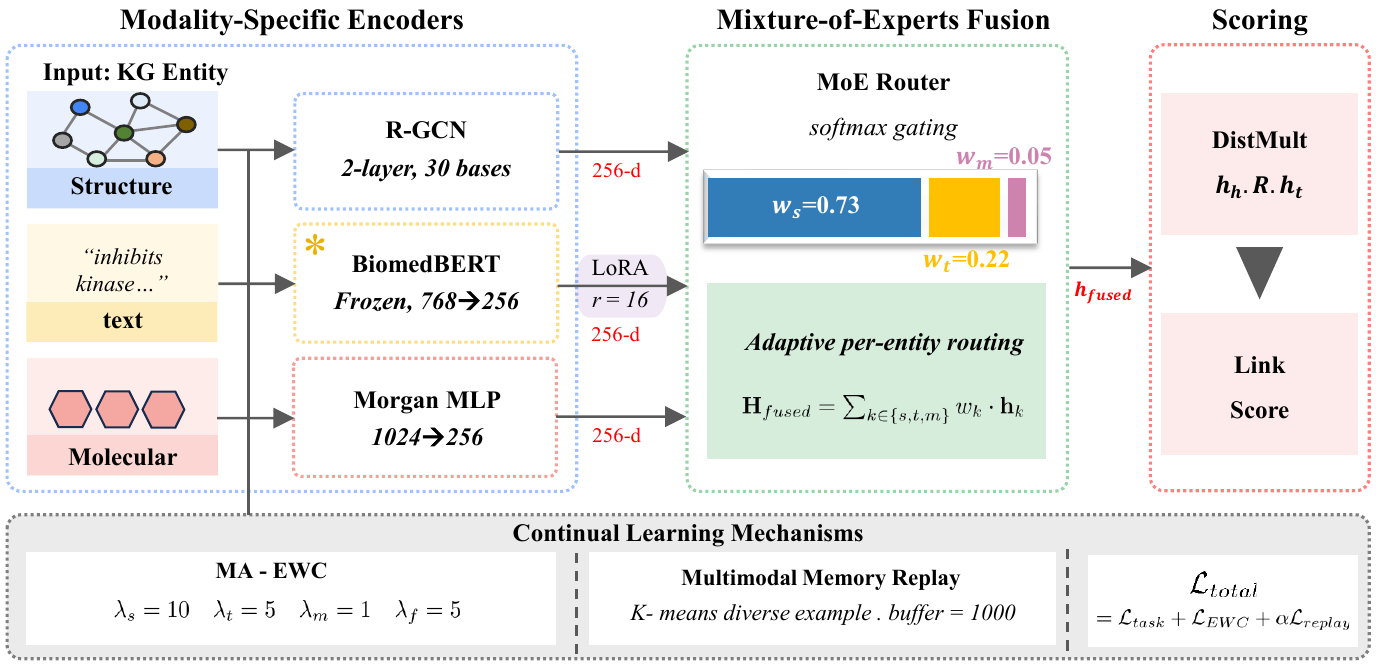} 
\caption{Architecture of \ours. Three modality-specific encoders produce structural ($\mathbf{h}^s$) via a 2-layer R-GCN, textual ($\mathbf{h}^t$) via frozen BiomedBERT with an optional LoRA adapter, and molecular ($\mathbf{h}^m$) via a Morgan-fingerprint MLP. A Mixture-of-Experts router adaptively combines them per entity via learned softmax weights ($w^s, w^t, w^m$). The fused embedding $\mathbf{h}_{\text{fused}}$ is scored by a DistMult decoder. Continual-learning mechanisms (bottom): modality-aware EWC applies per-modality Fisher regularization ($\lambda_s, \lambda_t, \lambda_m, \lambda_f$) and K-means multimodal memory replay selects diverse exemplars from past tasks into a size-1000 buffer. The total training loss combines task, EWC, and replay terms.}
\label{fig:architecture}
\end{figure*}

\ours consists of four stages: (i)~three modality-specific encoders that extract structural, textual, and molecular representations; (ii)~a Mixture-of-Experts (MoE) fusion module that adaptively combines per-modality representations via a learned router; (iii)~a decoder (DistMult) that scores candidate triples using the fused entity embeddings; and (iv)~continual learning mechanisms (standard EWC regularization and a multimodal memory replay buffer) that protect previously learned knowledge as the model trains on new tasks.

\subsection{Modality-Specific Encoders}
\label{subsec:encoders}

\paragraph{Structural ($\mathbf{h}_v^s$).} A 2-layer R-GCN~\citep{schlichtkrull2018modeling} with basis decomposition,
\begin{equation}
    \mathbf{h}_v^{(l+1)} = \sigma\!\left( \sum_{r \in \mathcal{R}} \sum_{u \in \mathcal{N}_v^r} \tfrac{1}{|\mathcal{N}_v^r|} \mathbf{W}_r^{(l)} \mathbf{h}_u^{(l)} + \mathbf{W}_0^{(l)} \mathbf{h}_v^{(l)} \right),
    \label{eq:rgcn}
\end{equation}
with $\mathbf{W}_r^{(l)} = \sum_{b=1}^{B} a_{rb}^{(l)} \mathbf{V}_b^{(l)}$ and ReLU $\sigma$, producing $\mathbf{h}_v^s\in\mathbb{R}^D$.

\paragraph{Textual ($\mathbf{h}_v^t$).} Frozen BiomedBERT~\citep{gu2021domain} mean-pooled and linearly projected, $\mathbf{h}_v^t = \mathbf{W}_{\text{proj}} \cdot \text{MeanPool}(\text{BiomedBERT}(x_v^{\text{text}})) + \mathbf{b}_{\text{proj}}$ with $\mathbf{W}_{\text{proj}}\in\mathbb{R}^{D\times768}$. Only the projection trains; entities without text get a learned default.

\paragraph{Molecular ($\mathbf{h}_v^m$).} For drug nodes, 1024-bit radius-2 Morgan fingerprints~\citep{rogers2010extended} are passed through a 2-layer MLP $\mathbf{h}_v^m = \mathbf{W}_2\,\sigma(\mathbf{W}_1 x_v^{\text{mol}}+\mathbf{b}_1)+\mathbf{b}_2$ with $\mathbf{W}_1\in\mathbb{R}^{D\times1024}$, $\mathbf{W}_2\in\mathbb{R}^{D\times D}$. Non-drug entities get a learned default.

\subsection{Multimodal Fusion}
\label{subsec:fusion}

The fusion module must satisfy three constraints imposed by the encoder asymmetry from \cref{sec:intro}: it must (i) be able to place all weight on a single modality when the data justify it, (ii) not push frozen-encoder outputs through a learned bottleneck that destroys their pretrained geometry, and (iii) allow post-hoc inspection of which modality each entity actually used. Among the embedding-level and score-level alternatives we evaluate (\cref{subsec:fusion_ablation}), Mixture-of-Experts (MoE) routing is the only one that satisfies all three.

\paragraph{Mixture-of-Experts Fusion (selected).}
We treat each modality encoder as an expert and learn a per-entity router that produces modality weights:
\begin{equation}
    \mathbf{w}_v = \text{softmax}\left(\mathbf{W}_r [\mathbf{h}_v^s; \mathbf{h}_v^t; \mathbf{h}_v^m] + \mathbf{b}_r\right) \in \mathbb{R}^3,
    \label{eq:moe_router}
\end{equation}
where $\mathbf{W}_r \in \mathbb{R}^{3 \times 3D}$ is the router weight matrix.
The fused representation is a weighted sum of the expert outputs:
\begin{equation}
    \mathbf{h}_v = w_v^s \cdot \mathbf{h}_v^s + w_v^t \cdot \mathbf{h}_v^t + w_v^m \cdot \mathbf{h}_v^m.
    \label{eq:moe_fusion}
\end{equation}
The router can assign weight $w_v^i = 1$ to any single expert, guaranteeing that the fused representation is at least as good as the best single modality. This property is central to handling the greedy modality problem (\cref{subsec:analysis}): when frozen text embeddings are unhelpful for training-loss reduction, the router suppresses them without destroying information through a bottleneck projection.

\paragraph{Alternative fusion variants.}
We compare MoE against gated cross-modal attention (4-head pairwise attention with an MLP and residual fuse), direct concatenation followed by projection at the embedding level, and score-level fusion with per-modality decoders~\citep{zhao2022mose}. Detailed numbers for all four variants appear in \cref{tab:fusion_ablation} and equations in \cref{sec:supp_fusion_eq}: embedding-level MoE/concat/gated cluster around AP $0.060$--$0.063$, while score-level fusion falls to $0.049$ because frozen BiomedBERT scores cannot serve as useful margin-ranking targets when the encoder cannot adapt.

\subsection{EWC Regularization (Per-Modality Configuration)}
\label{subsec:maewc}

We apply standard EWC~\citep{kirkpatrick2017overcoming} with a per-encoder $\lambda_k$ configuration that we frame as a tuning choice rather than a methodological contribution. Partitioning parameters by encoder, $\theta=\{\theta_s,\theta_t,\theta_m,\theta_f\}$, the regularizer is
\begin{equation}
    \mathcal{L}_{\text{EWC}} = \sum_{k \in \{s,t,m,f\}} \lambda_k \sum_i F_i^k \left(\theta_{k,i} - \theta_{k,i}^*\right)^2,
    \label{eq:maewc}
\end{equation}
with $\lambda_s{=}10,\lambda_t{=}5,\lambda_m{=}1,\lambda_f{=}5$ chosen on validation. Fisher matrices $F_i^k$ are diagonal and computed from squared gradients post-task. The ablation in \cref{subsec:cl_ablation} shows that per-encoder $\lambda$ matches uniform $\lambda$ on DistMult; we therefore treat per-encoder $\lambda$ as a tuning knob and retain it because it is conceptually well-defined per modality and may help on decoders where EWC has more headroom.

\subsection{Multimodal Memory Replay}
\label{subsec:replay}

\ours maintains a fixed-size buffer $\mathcal{M}$ of $|\mathcal{M}|$ representative triples balanced across all completed tasks. After each task $T_k$, we re-cluster: K-means is run on the structural embeddings of head entities from all triples currently in $\mathcal{M}\cup T_k$ with $K=|\mathcal{M}|/k$ clusters \emph{per task}, and the closest-to-centroid triple from each cluster is retained. This rebalancing strategy preserves diverse exemplars from \emph{every} prior task (at least $|\mathcal{M}|/k$ per task after task $k$) rather than purging early tasks via a strict FIFO queue. We cluster on structural embeddings because they cover $100\%$ of entities, whereas text and molecular features cover only $36\%$ and $6\%$, respectively; concatenating partial-coverage features would produce heterogeneous, high-dimensional cluster inputs. Selected triples are \emph{replayed with all available modalities} during subsequent tasks, which is what makes the buffer multimodal. Replay batches $\mathcal{B}_r$ are sampled uniformly from $\mathcal{M}$ and mixed with the current task, with loss
\begin{equation}
    \mathcal{L}_{\text{replay}} = \tfrac{1}{|\mathcal{B}_r|}\!\!\sum_{(h,r,t) \in \mathcal{B}_r}\!\!\ell(f_r(\mathbf{h}_h, \mathbf{h}_t), y).
    \label{eq:replay}
\end{equation}
\subsection{Training Procedure}
\label{subsec:training}

For each task $T_k$, all entities are encoded by the three modality-specific encoders, fused by the MoE router, and scored by DistMult. The total objective combines the task loss with EWC regularization and replay loss, the latter two active only for $k>1$:
\begin{equation}
    \mathcal{L} = \mathcal{L}_{\text{task}} + \mathcal{L}_{\text{EWC}} + \alpha \cdot \mathcal{L}_{\text{replay}}.
    \label{eq:total}
\end{equation}
After each task, we compute per-modality Fisher matrices and add K-means-selected diverse exemplars to $\mathcal{M}$. Full pseudocode appears in \cref{sec:supp_alg}.

\section{Experiments}
\label{sec:experiments}
\subsection{Experimental Setup}
\label{subsec:setup}
\paragraph{Benchmark.}
We evaluate on a continual learning benchmark derived from PrimeKG~\citep{chandak2023building} with two temporal snapshots ($t_0$: June 2021; $t_1$: July 2023) reconstructed from nine biomedical databases. The benchmark contains 129K+ entities across 10 node types and 8.1M+ edges spanning 30 relation types, organized into 10 continual learning tasks via entity-type grouping. Within each task, triples are split into train, validation, and test at a 70/10/20 ratio. We follow standard CL protocol~\citep{lopez2017gradient, chaudhry2019efficient} and report Average Performance (AP, the mean filtered MRR across all tasks after the final training step), Average Forgetting (AF), Backward Transfer (BWT), and Remembering (REM). Formal definitions are released with the benchmark.
\paragraph{Baselines and metrics.}
We compare against Naive Sequential, Joint Training, EWC~\citep{kirkpatrick2017overcoming}, and LKGE~\citep{cui2023lifelong}. All baselines share the DistMult decoder (dim 256) with \ours for a fair comparison, except LKGE, which requires TransE. AP for relationship prediction uses filtered MRR over 129K+ candidate entities; AP for entity classification uses Macro-F1.
\paragraph{Implementation details.}
\ours uses embedding dimension $D{=}256$, a 2-layer R-GCN with $B{=}30$ basis matrices, frozen BiomedBERT (microsoft/BiomedNLP-BiomedBERT-base-uncased-abstract) with a $768{\to}256$ linear projection, a Morgan-fingerprint MLP ($1024{\to}256{\to}256$), MoE fusion with a 3-way softmax router, and a DistMult decoder~\citep{yang2015embedding}. Training uses Adam at learning rate $10^{-3}$ for 100 epochs per task, batch size $512$, and a memory buffer of size $1000$. EWC strengths are $\lambda_s{=}10$, $\lambda_t{=}5$, $\lambda_m{=}1$, $\lambda_f{=}5$, with replay weight $\alpha{=}1.0$. All experiments use $5$ random seeds $\{42, 123, 456, 789, 1024\}$ on NVIDIA V100 GPUs ($32$\,GB) and report mean $\pm$ std.

\subsection{Biomedical Relationship Prediction (Link Prediction) Results}
\label{subsec:main_results}

\begin{table}[t]
\caption{\ours on continual biomedical relationship prediction (10 tasks, 5 seeds, DistMult). AP: filtered MRR over 129K+ entities; baselines share the DistMult decoder for fair comparison.}
\label{tab:main_results}
\centering
\small
\setlength{\tabcolsep}{4pt}
\begin{tabular}{lcccc}
\toprule
\textbf{Variant} & \textbf{AP} $\uparrow$ & \textbf{AF} $\downarrow$ & \textbf{BWT} $\uparrow$ & \textbf{REM} $\uparrow$ \\
\midrule
\rowcolor{blond} \ours (MoE) & $0.062 \pm 0.010$ & $0.043 \pm 0.007$ & $-0.043 \pm 0.007$ & $0.957 \pm 0.007$ \\
\quad w/o text \& mol (struct only) & $0.071 \pm 0.005$ & $0.043 \pm 0.005$ & $-0.043 \pm 0.005$ & $0.957 \pm 0.005$ \\
\quad w/o struct \& mol (text only) & $\mathbf{0.136 \pm 0.003}$ & $0.007 \pm 0.001$ & $-0.007 \pm 0.001$ & $0.993 \pm 0.001$ \\
\midrule
Naive Sequential & $0.058 \pm 0.001$ & $0.004 \pm 0.001$ & $-0.004 \pm 0.001$ & $0.996 \pm 0.001$ \\
EWC~\citep{kirkpatrick2017overcoming} & $0.058 \pm 0.001$ & $0.006 \pm 0.001$ & $-0.006 \pm 0.001$ & $0.994 \pm 0.001$ \\
Joint Training & $0.047 \pm 0.001$ & --- & --- & --- \\
LKGE~\citep{cui2023lifelong} (TransE) & $0.039 \pm 0.001$ & $0.012 \pm 0.002$ & $-0.010 \pm 0.003$ & $0.990 \pm 0.003$ \\
\bottomrule
\end{tabular}
\end{table}

\begin{figure}[t]
\centering
\includegraphics[width=\linewidth]{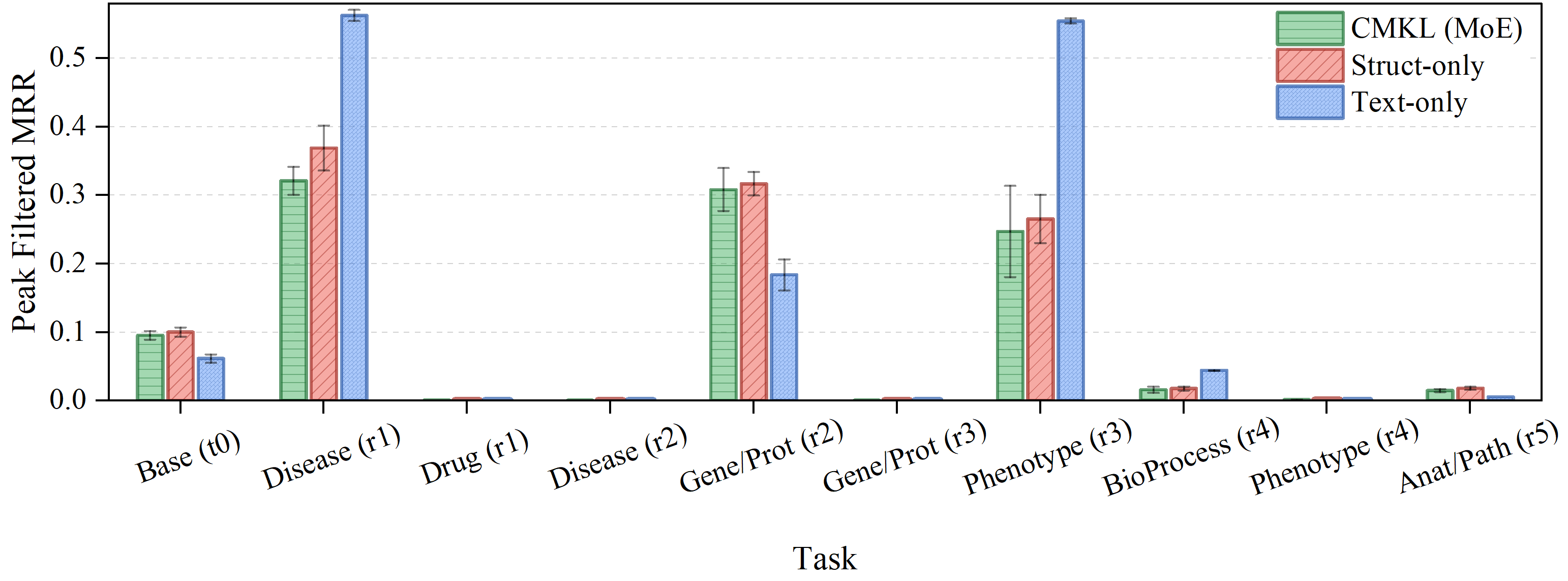}
\caption{Per-task peak MRR for \ours vs.\ single-modality ablations. Text-only excels on text-rich tasks (Disease-r1, Phenotype-r3); structural features drive other tasks; greedy-modality dynamics analyzed in \cref{subsec:analysis}.}
\label{fig:per_task}
\end{figure}

\cref{tab:main_results} reports \ours's performance on continual biomedical relationship prediction, alongside modality ablations. \ours reaches AP\,=\,$0.062 \pm 0.010$, the highest among the jointly trained multimodal methods. Same-decoder comparisons separate cleanly: \ours outperforms Joint Training ($0.047$) by $1.3\times$ (paired $t$-test, $p{=}0.045$) and LKGE ($0.039$, TransE-constrained) by $1.6\times$. The structural-only ablation reaches a higher AP ($0.071$, see ``modality ablations'' rows of \cref{tab:main_results}), which is a key finding we analyze in \cref{subsec:analysis}: jointly training a frozen text encoder with a learnable structural one trades raw LP performance for stable multimodal representations that drive the entity-classification gain in \cref{subsec:nc_results}. The nominal $7\%$ gap over Naive Sequential ($0.058 \pm 0.001$) and EWC ($0.058 \pm 0.001$) is not statistically significant ($p{=}0.48$ vs.\ Naive, $p{=}0.40$ vs.\ EWC), because \ours's seed-to-seed standard deviation ($0.010$) exceeds the gap. The 95\% bootstrap confidence intervals (10K resamples over five seeds) are $[0.054, 0.071]$ for \ours, $[0.057, 0.059]$ for Naive, and $[0.057, 0.058]$ for EWC, all overlapping and consistent with the paired-test result. We therefore read \ours's link-prediction contribution as matching the strongest structural CL methods within noise, while delivering markedly different per-task and per-stratum behavior (\cref{fig:per_task}, \cref{fig:modality_forgetting}); the materially significant multimodal gain appears on entity classification ($+60\%$, \S\ref{subsec:nc_results}).

\paragraph{Frozen text as an evaluation-time ceiling.}
The text-only ablation reaches AP $0.136$, double full \ours ($0.062$) and nearly double structural-only ($0.071$). Frozen BiomedBERT receives no link-prediction training signal yet encodes semantic similarity between entity descriptions, and on filtered MRR over 129K candidates this pretrained clustering is itself a strong ranker. However, this signal is \emph{unreachable by margin-ranking gradients}: the BiomedBERT encoder is frozen, so the loss cannot reshape its features; only the linear projection layer on top is trained, and the loss surface induced by margin ranking is not aligned with cosine-style semantic geometry, so a fully trainable encoder optimizing against the same objective does not converge to the same solution. The text-only ablation is therefore an evaluation-time ceiling on the encoder side: it carries small per-task drift through the trainable projection and decoder (AF $\approx 0.007$ in \cref{tab:main_results}), but the underlying representation that produces the $0.136$ AP is encoder-level and inaccessible to the trainable parts. Any joint fusion that exposes the frozen encoder to gradients indirectly suppresses its contribution within a few epochs (\cref{subsec:analysis}). The $0.136$ figure indicates where the encoder-level headroom is, $0.062$ shows how much of it a trainable continual learner currently recovers, and the gap is the cost of the asymmetry MoE routing is designed to manage.

\subsection{Biomedical Entity Classification (Node Classification) Results}
\label{subsec:nc_results}

\begin{table}[t]
\caption{\ours on continual biomedical entity classification (10 tasks, 5 seeds). AP: mean Macro-F1.}
\label{tab:nc_results}
\centering
\small
\begin{tabular}{lccc}
\toprule
\textbf{Variant} & \textbf{AP} $\uparrow$ & \textbf{AF} $\downarrow$ & \textbf{BWT} $\uparrow$ \\
\midrule
\rowcolor{blond} \ours (MoE, all modalities) & $\mathbf{0.591 \pm 0.005}$ & $\mathbf{0.008 \pm 0.009}$ & $\mathbf{+0.003 \pm 0.007}$ \\
\quad w/o text \& mol (struct only) & $0.345 \pm 0.004$ & $0.011 \pm 0.002$ & $+0.003 \pm 0.005$ \\
\midrule
Best baseline (Joint Training) & $0.370 \pm 0.002$ & --- & --- \\
\bottomrule
\end{tabular}
\end{table}

\cref{tab:nc_results} reports results on continual biomedical entity classification, where the model must assign the correct biomedical type (gene/protein, disease, drug, etc.) to each entity as the graph evolves. \ours reaches AP\,=\,$0.591 \pm 0.006$, outperforming the strongest structural-only baseline (Joint Training, $0.370$) by $60\%$ ($+0.221$ absolute), with near-zero forgetting (AF\,=\,$0.008$) and positive backward transfer (BWT\,=\,$+0.003$).

The source of this gap deserves explicit attribution. \ours's \emph{structural-only} ablation (AP\,=\,$0.345$, \cref{tab:nc_results} row 2) lies \emph{below} Joint Training, so the $0.221$ gain is a multimodal-features effect, not a CMKL-architecture effect: BiomedBERT's textual descriptions are strongly type-discriminative, and any classifier with access to them would close most of the gap. \ours's contribution on this track is therefore to \emph{preserve the multimodal-feature advantage across the continual sequence} (AF near zero, BWT positive), not to extract a signal a simpler multimodal classifier could not. A ``Joint + multimodal'' oracle would be a fair feature-availability control for future comparison, but lies outside the continual-evolution focus of this paper.
\subsection{Fusion Strategy Ablation}
\label{subsec:fusion_ablation}

\begin{figure}[t]
\centering
\includegraphics[width=\linewidth]{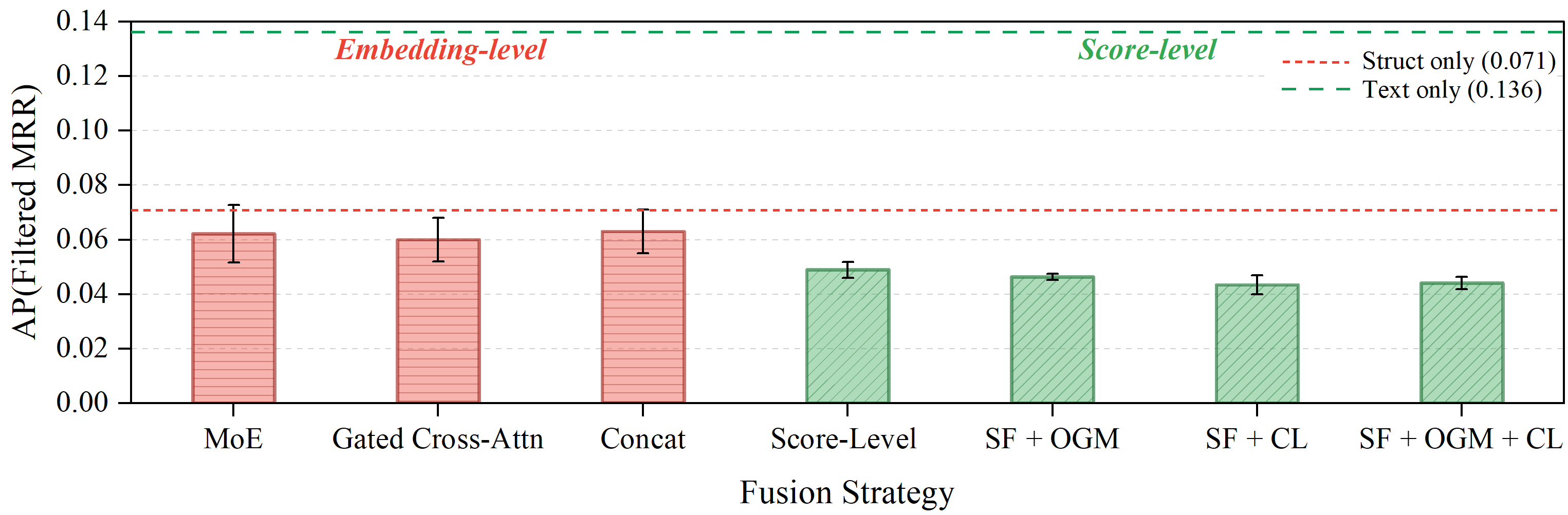}
\caption{Fusion-strategy comparison. Embedding-level (MoE, gated, concat) outperforms score-level. Dashed lines: single-modality references.}
\label{fig:fusion_bar}
\end{figure}

\cref{fig:fusion_bar} and \cref{tab:fusion_ablation} compare seven fusion strategies (three embedding-level and four score-level) on the same DistMult + EWC backbone. Among the embedding-level strategies, concatenation and MoE reach comparable AP ($0.063$ and $0.062$), with concatenation achieving the lowest forgetting (AF\,=\,$0.024$). We select MoE as the primary fusion because its learned router provides interpretable modality weights that expose the greedy-modality dynamics analyzed in \cref{subsec:analysis}; concatenation hides the per-entity modality preference inside its $3D{\to}D$ projection. All score-level variants underperform the embedding-level approaches, with the best score-level result (SF-only, AP\,=\,$0.049$) falling $21\%$ below MoE.

\begin{table}[t]
\caption{Fusion strategy comparison on continual biomedical relationship prediction (5 seeds, DistMult+EWC). MoE, gated cross-attention, and concatenation operate at the embedding level; score-level (SF) variants use per-modality decoders, optionally combined with OGM-GE gradient modulation~\citep{peng2022balanced} or contrastive alignment.}
\label{tab:fusion_ablation}
\centering
\small
\begin{tabular}{lcc}
\toprule
\textbf{Fusion Strategy} & \textbf{AP} $\uparrow$ & \textbf{AF} $\downarrow$ \\
\midrule
\textbf{MoE (selected)}     & $\mathbf{0.062 \pm 0.010}$ & $0.043 \pm 0.007$ \\
Gated Cross-Attention       & $0.060 \pm 0.008$           & $0.039 \pm 0.007$ \\
Concatenation               & $0.063 \pm 0.008$           & $\mathbf{0.024 \pm 0.009}$ \\
\midrule
Score-Level (SF-only)       & $0.049 \pm 0.003$           & $0.039 \pm 0.002$ \\
SF + OGM-GE                 & $0.046 \pm 0.001$           & $0.035 \pm 0.005$ \\
SF + Contrastive            & $0.043 \pm 0.003$           & $0.038 \pm 0.007$ \\
SF + OGM + Contrastive      & $0.044 \pm 0.002$           & $0.040 \pm 0.001$ \\
\bottomrule
\end{tabular}
\end{table}

\paragraph{Why score-level fusion underperforms.}
Score-level fusion~\citep{zhao2022mose} trains per-modality decoders independently to decouple training signals. The structural decoder learns useful ranking patterns from R-GCN outputs, but the text decoder cannot: margin-ranking gradients reshape decoder weights yet cannot reshape an encoder that does not move. Adding OGM-GE gradient modulation (AP\,=\,$0.046$) provides no benefit, as the modulation weights remain $\approx\!1$ throughout training; the structural and text decoders converge at similar speeds because the text decoder is learning nothing useful, not being suppressed. An $\alpha$-sweep over the text decoder weight confirms that lower text contribution is better ($\alpha{=}0.3$: $0.059$ vs.\ $\alpha{=}2.0$: $0.035$). This is the same encoder-level asymmetry the MoE router resolves by suppression rather than by re-weighting decoders that cannot themselves recover it.

\subsection{Continual Learning Component Ablation}
\label{subsec:cl_ablation}
\cref{tab:cl_ablation} factors the contribution of the two CL mechanisms in CMKL: per-encoder EWC and the K-means-diverse multimodal replay buffer. Three findings stand out. (a) On DistMult, per-encoder $\lambda$ matches uniform $\lambda{=}10$ within seed noise (both at AP $\approx 0.063$, AF $\approx 0.043$); we therefore treat per-encoder $\lambda$ as a tuning knob rather than a methodological contribution, while retaining it because it is conceptually well-defined per modality and may matter on decoders where EWC has more headroom. (b) Removing EWC entirely ($\lambda{=}0$) preserves AP ($0.064$) but increases AF to $0.050$, indicating that the replay buffer is the primary forgetting-mitigation mechanism at this scale and EWC provides a marginal AF reduction. (c) Performance is robust to a $50\times$ range of $\lambda$ ($0.1$--$5\times$, AP $0.059$--$0.065$) and to buffer sizes $500$--$5{,}000$ (AP $0.060$--$0.066$); practitioners need not tune these carefully on this benchmark, and even the largest buffer covers $\approx 0.06\%$ of the $8.1$M edges, making replay information-cheap relative to the graph size.

\begin{table}[t]
\caption{Continual learning component ablation (5 seeds, MoE-DistMult). (a) EWC strategy. (b) $\lambda$ sensitivity (per-encoder, scaled by a common factor relative to default). (c) Replay buffer size.}
\label{tab:cl_ablation}
\centering
\small
\setlength{\tabcolsep}{4pt}
\begin{tabular}{lcc}
\toprule
\textbf{Configuration} & \textbf{AP} $\uparrow$ & \textbf{AF} $\downarrow$ \\
\midrule
\multicolumn{3}{l}{\emph{(a) EWC strategy}} \\
Per-encoder EWC (default)      & $0.062 \pm 0.010$ & $0.043 \pm 0.007$ \\
Uniform EWC ($\lambda{=}10$)   & $0.063 \pm 0.003$ & $0.043 \pm 0.002$ \\
No EWC ($\lambda{=}0$)         & $0.064 \pm 0.006$ & $0.050 \pm 0.009$ \\
\midrule
\multicolumn{3}{l}{\emph{(b) $\lambda$ sensitivity (relative to default)}} \\
$0.1\times$                    & $0.059 \pm 0.007$ & $\mathbf{0.039 \pm 0.001}$ \\
$0.5\times$                    & $0.064 \pm 0.003$ & $0.040 \pm 0.003$ \\
$1\times$ (default)            & $0.062 \pm 0.010$ & $0.043 \pm 0.007$ \\
$2\times$                      & $0.064 \pm 0.011$ & $0.044 \pm 0.007$ \\
$5\times$                      & $\mathbf{0.065 \pm 0.005}$ & $0.042 \pm 0.006$ \\
\midrule
\multicolumn{3}{l}{\emph{(c) Replay buffer size}} \\
$500$                          & $0.060 \pm 0.005$ & $\mathbf{0.041 \pm 0.009}$ \\
$1{,}000$ (default)            & $0.062 \pm 0.010$ & $0.043 \pm 0.007$ \\
$2{,}000$                      & $\mathbf{0.066 \pm 0.007}$ & $0.049 \pm 0.001$ \\
$5{,}000$                      & $0.062 \pm 0.004$ & $0.048 \pm 0.007$ \\
\bottomrule
\end{tabular}
\end{table}

\subsection{Analysis}
\label{subsec:analysis}

\paragraph{AF must be read against per-task peak MRR.}
\ours's AF $=0.043$ exceeds EWC's $0.006$, but the comparison is misleading without context. Naive Sequential and EWC peak below $0.1$ MRR on most tasks, leaving almost no signal to lose, while \ours peaks at $0.32$ on Disease-r1, so its drift is measured against a much taller curve. \ours's Remembering score $=0.957$ retains $95.7\%$ of acquired knowledge in absolute terms, and Naive/EWC's lower AF reflects flatter learning rather than stronger retention; an aggressive regularization strategy could trivially achieve AF $\approx 0$ by capping AP near $0.03$. \ours sits on the efficient corner of the AP--AF frontier.

\begin{figure}[t]
\centering
\includegraphics[width=\linewidth]{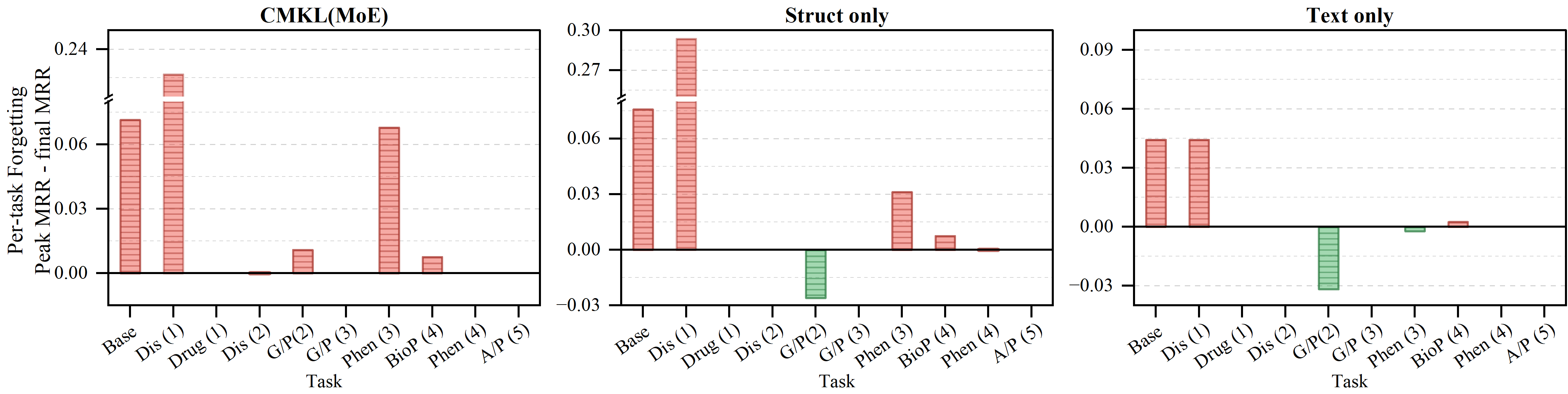}
\caption{Per-task forgetting (peak $-$ final MRR) by modality. Red: positive; green: backward transfer. Structural encoder dominates forgetting; frozen text is inherently stable.}
\label{fig:modality_forgetting}
\end{figure}

\paragraph{Modality-specific forgetting and the greedy-modality problem.}
\cref{fig:modality_forgetting} shows per-task forgetting by modality. The structural-only ablation forgets sharply on Disease-r1 ($0.29$) and Base-$t_0$ ($0.08$); the text-only ablation is uniformly stable ($\leq 0.05$, frozen by construction); the full \ours model shifts the dominant source of forgetting to Disease-r1 ($0.23$). The MoE router assigns $w^s \approx 0.70$--$1.00$ and suppresses text ($w^t \approx 0.00$--$0.25$), reflecting a known convergence asymmetry between learnable and frozen encoders: the R-GCN reduces margin loss faster than frozen BiomedBERT~\citep{wu2022characterizing}. This localizes the greedy-modality problem at the \emph{representation level}, not the fusion level: gradient modulation in the fusion stage cannot fix what the encoder freeze causes, while MoE routing can suppress an unreachable modality without bottlenecking it.

\paragraph{Scope: bilinear-decoder family.}
\ours is deliberately scoped to the bilinear (DistMult) decoder family. A real-valued additive MoE over real-valued encoder outputs composes cleanly with bilinear scoring, but not with rotational or Hermitian scoring. We attempted ComplEx and RotatE integrations: across four successive design iterations, \ours-RotatE plateaus at AP $0.025 \pm 0.003$ (vs.\ baseline EWC-RotatE $0.088$) and \ours-ComplEx at AP $0.021 \pm 0.005$ (vs.\ baseline EWC-ComplEx $0.029$). The failure mode is geometric rather than statistical: a linear fusion of real-valued encoder outputs does not respect the rotational metric a RotatE decoder is built to exploit, and the gap appears at task-0 initial training (CMKL-RotatE task-0 MRR $\approx 0.12$ vs.\ baseline RotatE $\approx 0.26$) rather than as a forgetting effect. We frame this as a precise statement of where the recipe currently applies, and as a concrete design problem for future work: a decoder-aware fusion such as complex-valued routing or near-identity initialization in the rotational metric would extend the same machinery to those families.

\section{Conclusion}
\label{sec:conclusion}

We presented \fullname (\ours), a continual KG learning framework that fuses structural (R-GCN), textual (frozen BiomedBERT), and molecular (Morgan MLP) encoders via MoE routing, and protects previously learned knowledge via EWC and K-means-diverse multimodal replay. On a 129K-entity benchmark, \ours matches the strongest structural CL methods on biomedical relationship prediction (AP\,=\,$0.062\pm0.010$, vs.\ Naive/EWC $0.058$ within seed noise; significantly above LKGE $0.039$ and Joint Training $0.047$, $p{=}0.045$) and preserves the multimodal-feature $+60\%$ entity-classification advantage across the sequence with near-zero forgetting ($0.591$ vs.\ Joint $0.370$). The central methodological lesson is that the greedy modality problem is encoder-level, not fusion-level: a frozen text encoder posts a higher per-task ceiling than any jointly trained model (text-only AP $0.136$) but cannot be reached by margin-ranking gradients, and any fusion that exposes it to those gradients suppresses it within a few epochs. MoE routing addresses this asymmetry by allowing the router to suppress an unreachable modality without forcing it through a learned bottleneck; we adopt it primarily for its diagnostic and interpretability value, with concatenation a comparable alternative when interpretability is not required (\cref{tab:fusion_ablation}).

\paragraph{Limitations and future work.} \emph{(1) LP significance.} The LP gap over Naive/EWC is not statistically significant ($p{=}0.48,\, 0.40$); we read CMKL as matching the strongest structural baselines with different per-task/per-stratum behavior rather than uniform gains. \emph{(2) NC gain source.} The 60\% NC headline is feature-driven (struct-only ablation $<$ Joint), so CMKL's contribution is preserving the multimodal-feature advantage across the sequence rather than producing it. \emph{(3) Decoder scope.} \ours is scoped to bilinear (DistMult) decoders: a real-valued additive MoE does not respect rotational or Hermitian geometry, and our extension attempts plateaued well below the corresponding RotatE/ComplEx baselines (\cref{subsec:analysis}); a decoder-aware fusion (complex-valued routing, near-identity init in the rotational metric) is the natural follow-up.

\bibliographystyle{unsrt}
{\small
\bibliography{refs}
}

\newpage
\setcounter{page}{1}
\renewcommand{\thefigure}{S\arabic{figure}}
\setcounter{figure}{0}
\renewcommand{\thesection}{S\arabic{section}}
\setcounter{section}{0}
\renewcommand{\thetable}{S\arabic{table}}
\setcounter{table}{0}

\begin{center}
\Large\textbf{CMKL: Modality-Aware Continual Learning for Evolving Biomedical Knowledge Graphs}\par
\vspace{1em}
\Large Supplementary Material
\end{center}

\section{Experimental Details}
\label{sec:supp_exp}
All experiments use NVIDIA V100 GPUs (32\,GB) with PyTorch 2.0 and PyTorch Geometric. Training uses Adam optimizer with learning rate 0.001, batch size 512, and 50{,}000 sampled triples per epoch for 100 epochs per task. Embedding dimension: 256. R-GCN: 2 layers, 30 basis matrices. MoE router: hidden dimension 128, load-balance weight 0.01. MA-EWC lambdas: $\lambda_s = 10$, $\lambda_t = 5$, $\lambda_m = 1$, $\lambda_f = 5$. Replay buffer: 1{,}000 total triples, K-means selection on structural embeddings. Score-level fusion: $\alpha_{\text{text}} = 0.5$, $\alpha_{\text{mol}} = 0.3$. Contrastive alignment: temperature 0.1, weight 0.1. OGM-GE: $\alpha = 1.0$. All results averaged over 5 seeds $\{42, 123, 456, 789, 1024\}$.

\section{Additional Results}
\label{sec:supp_results}
\paragraph{Alpha sensitivity.} The score-fusion text weight ($\alpha_{\text{text}}$) sweep (single seed, seed=42) shows: $\alpha=0.3$ gives AP=0.059 (best), $\alpha=0.5$ gives 0.047, $\alpha=1.0$ gives 0.049, $\alpha=2.0$ gives 0.035 (worst). Lower text weight is consistently better, confirming the dominance of structural features in biomedical relationship prediction.

\paragraph{Forward Transfer.} FWT equals zero for all methods in our sequential setting, as tasks are non-overlapping and models are not evaluated on future tasks before training.

\section{Limitations and Future Work}
\label{sec:supp_lim}
\begin{itemize}
\item \textbf{Frozen-vs-learnable encoder asymmetry.} The text-only ablation (AP\,=\,$0.136$, \cref{tab:main_results}) outperforms fusion-based methods on link prediction because frozen BiomedBERT encodes evaluation-time semantic similarity that margin-based training cannot access. A natural future direction is parameter-efficient text-encoder fine-tuning (e.g., LoRA adapters); this would let the textual encoder drift slightly under task signal while preserving most of its pretrained geometry.
\item \textbf{MoE router collapse.} The router occasionally collapses to a single expert (structural weight $w^s{=}1.0$ on one seed), suggesting training instability that load-balancing losses could mitigate.
\item \textbf{MA-EWC vs uniform EWC.} On DistMult, MA-EWC shows no advantage over uniform EWC; per-modality variants may matter more on decoders where EWC's effect is larger.
\item \textbf{Bilinear-decoder scope.} \ours is designed around DistMult. Extending the multimodal + MA-EWC recipe to complex-valued decoders (ComplEx, RotatE) requires a fusion architecture that respects their geometric constraints.
\item \textbf{Modality coverage.} Text covers 36\% of entities and molecular features 6\%, bounding the potential contribution of those modalities.
\item \textbf{Other future directions} include extending evaluation to additional temporal snapshots, modality-specific learning-rate schedules, and direct per-encoder Fisher analysis.
\end{itemize}

\section{Fusion-Strategy Equations}
\label{sec:supp_fusion_eq}

The two embedding-level variants summarized in the main paper are:

\paragraph{Gated cross-modal attention.}
For each entity $v$ we form a 3-token modality sequence $\mathbf{S}_v = [\mathbf{h}_v^s;\mathbf{h}_v^t;\mathbf{h}_v^m] \in \mathbb{R}^{3\times D}$ and apply $H{=}4$-head pairwise multi-head attention across modality pairs $(i,j)$:
\begin{equation*}
\text{CrossAttn}(\mathbf{S}_v^i, \mathbf{S}_v^j) = \text{softmax}\!\big(\mathbf{Q}_i \mathbf{K}_j^\top/\sqrt{D/H}\big)\mathbf{V}_j,
\end{equation*}
where $\mathbf{Q}_i = \mathbf{S}_v^i \mathbf{W}_Q^i$ and $\mathbf{K}_j,\mathbf{V}_j$ are computed analogously over the $j$-th modality token; the softmax is therefore over the per-head key dimension, not over a single scalar. Stacking the four cross-attended outputs together with the structural anchor $\mathbf{h}_v^s$ gives $\mathbf{c}_v=[\mathbf{h}_v^s;\hat{\mathbf{h}}_v^{s\leftarrow t};\hat{\mathbf{h}}_v^{s\leftarrow m};\hat{\mathbf{h}}_v^{t\leftarrow s};\hat{\mathbf{h}}_v^{m\leftarrow s}]\in\mathbb{R}^{5D}$ and a final MLP-with-residual fuse $\mathbf{h}_v=\text{LayerNorm}(\text{MLP}_{\text{fuse}}(\mathbf{c}_v)+\mathbf{h}_v^s)$. Structural anchoring (using $\mathbf{h}_v^s$ as the residual base) reflects that structural features cover all entities while text and molecular features cover only $36\%$ and $6\%$.

\paragraph{Concatenation.}
$\mathbf{h}_v=\mathbf{W}_c[\mathbf{h}_v^s;\mathbf{h}_v^t;\mathbf{h}_v^m]+\mathbf{b}_c$, $\mathbf{W}_c\in\mathbb{R}^{D\times3D}$.

\paragraph{Why score-level fusion underperforms.}
Score-level fusion~\citep{zhao2022mose} trains per-modality decoders to combine at evaluation time. Independent text decoders cannot learn useful margin-ranking targets from frozen BiomedBERT embeddings, so the modality mismatch reappears at the encoder level. OGM-GE gradient modulation~\citep{peng2022balanced} provided no benefit (modulation weights stayed $\approx1$); an $\alpha$ sweep over text-decoder weight confirmed less text is better ($\alpha=0.3$: AP\,=\,$0.059$ vs.\ $\alpha=2.0$: $0.035$).

\section{Training Algorithm}
\label{sec:supp_alg}

\begin{algorithm}[h]
\caption{\ours Training Procedure}
\label{alg:cmkl}
\begin{algorithmic}[1]
\REQUIRE Task sequence $\mathcal{T}=\{T_1,\ldots,T_K\}$, modality features $\mathbf{X}$, buffer size $|\mathcal{M}|$
\ENSURE Trained parameters $\theta=\{\theta_s,\theta_t,\theta_m,\theta_f\}$
\STATE Initialize encoders, MoE router, decoder; $\mathcal{M}\leftarrow\emptyset$
\FOR{$k=1$ \TO $K$}
    \FOR{each epoch}
        \STATE $\mathbf{H}^s\!\leftarrow\!\text{R-GCN}(\mathcal{V}_k,\mathcal{E}_k;\theta_s)$;\;$\mathbf{H}^t\!\leftarrow\!\text{BiomedBERT-Proj}(\mathbf{X}^{\text{text}};\theta_t)$;\;$\mathbf{H}^m\!\leftarrow\!\text{MLP}(\mathbf{X}^{\text{mol}};\theta_m)$
        \STATE $\mathbf{w}\leftarrow\text{softmax}(\mathbf{W}_r[\mathbf{H}^s;\mathbf{H}^t;\mathbf{H}^m]+\mathbf{b}_r)$;\;$\mathbf{H}\leftarrow w^s\mathbf{H}^s+w^t\mathbf{H}^t+w^m\mathbf{H}^m$
        \STATE $\mathcal{L}_{\text{task}}\leftarrow$ link-prediction loss on $T_k$ via decoder($\mathbf{H}$)
        \IF{$k>1$}
            \STATE $\mathcal{L}_{\text{EWC}}\leftarrow$ \cref{eq:maewc};\;$\mathcal{L}_{\text{replay}}\leftarrow$ replay loss on $\mathcal{B}_r\subset\mathcal{M}$
        \ELSE
            \STATE $\mathcal{L}_{\text{EWC}}\!\leftarrow\!0$;\;$\mathcal{L}_{\text{replay}}\!\leftarrow\!0$
        \ENDIF
        \STATE $\mathcal{L}\leftarrow\mathcal{L}_{\text{task}}+\mathcal{L}_{\text{EWC}}+\alpha\mathcal{L}_{\text{replay}}$;\;Adam update on $\theta$
    \ENDFOR
    \STATE Compute $\mathbf{F}^s,\mathbf{F}^t,\mathbf{F}^m,\mathbf{F}^f$;\;$\theta^*\leftarrow\theta$;\;$\mathcal{M}\leftarrow\text{K-means-Rebalance}(\mathcal{M}\cup T_k,\,\mathbf{H}^s,\,|\mathcal{M}|)$
\ENDFOR
\end{algorithmic}
\end{algorithm}

\section{Contextualizing Absolute MRR Values}
\label{sec:supp_mrr_context}
Filtered MRR over 129K+ candidate entities is inherently challenging. KG-FIT~\citep{jiang2024kgfit} reports TransE MRR\,=\,$0.048$ on a 10K-entity PrimeKG subset; scaling to 129K entities makes ranking proportionally harder. The meaningful comparisons in \cref{tab:main_results} are therefore relative differences between methods under the same decoder: \ours outperforms DistMult Joint Training by $1.3\times$ and LKGE (TransE-constrained) by $1.6\times$.

\section{Data Availability}
\label{sec:supp_data}
Our code and pretrained \ours models will be made publicly available on GitHub upon publication.


\end{document}